\documentclass[10pt,twocolumn,letterpaper]{article}

\usepackage{cvpr}              
\usepackage{colortbl}
\usepackage{enumitem}
\usepackage{stfloats}
\usepackage{xcolor}
\definecolor{peach}{RGB}{255, 180, 180}
\definecolor{orange}{RGB}{255, 220, 180}
\usepackage{multirow} 
\usepackage{graphicx}
\usepackage{subcaption}
\usepackage[ruled,vlined]{algorithm2e}
\usepackage[section]{placeins}
\usepackage{tabularx}
\usepackage{float}
\usepackage{lipsum}
\usepackage{enumitem}

\definecolor{cvprblue}{rgb}{0.21,0.49,0.74}
\usepackage[pagebackref,breaklinks,colorlinks,citecolor=cvprblue]{hyperref}

\def\paperID{39} 
\def\confName{3DV\xspace}
\def\confYear{2025\xspace}

\begin{document}
\title{Mipmap-GS: Let Gaussians Deform with Scale-specific Mipmap for Anti-aliasing Rendering}
\author{Jiameng Li$^{1}$~~~Yue Shi$^{2,3}$~~~Jiezhang Cao$^{2}$~~~Bingbing Ni$^{3}$~~~Wenjun Zhang$^{3}$ \\~~~Kai Zhang$^{4}$~~~Luc Van Gool$^{2,5}$
\\
$^{1}${University of Stuttgart}~~~$^{2}${ETH Z\"urich}~~~$^{3}${Shanghai Jiaotong University}\\~~~$^{4}${Nanjing University}~~~$^{5}${INSAIT, Sofia University}}
\maketitle
\begin{abstract}
3D Gaussian Splatting (3DGS) has attracted great attention in novel view synthesis because of its superior rendering efficiency and high fidelity. However, the trained Gaussians suffer from severe zooming degradation due to non-adjustable representation derived from single-scale training. Though some methods attempt to tackle this problem via post-processing techniques such as selective rendering or filtering techniques towards primitives, the scale-specific information is not involved in Gaussians. In this paper, we propose a unified optimization method to make Gaussians adaptive for arbitrary scales by self-adjusting the primitive properties (e.g., color, shape and size) and distribution (e.g., position). Inspired by the mipmap technique, we design pseudo ground-truth for the target scale and propose a scale-consistency guidance loss to inject scale information into 3D Gaussians. Our method is a plug-in module, applicable for any 3DGS models to solve the zoom-in and zoom-out aliasing. Extensive experiments demonstrate the effectiveness of our method. Notably, our method outperforms 3DGS in PSNR by an average of 9.25 dB for zoom-in and 10.40 dB for zoom-out on the NeRF Synthetic dataset. Our project website: \url{https://github.com/renaissanceee/Mipmap-GS}.
\end{abstract}



\vspace{-0.4cm}
\section{Introduction}
\label{sec:intro}

\begin{figure}[t]
    \centering
    \vspace{-1cm}
    \includegraphics[trim=9cm 7cm 8.3cm 6cm,clip=true,width=1\linewidth]{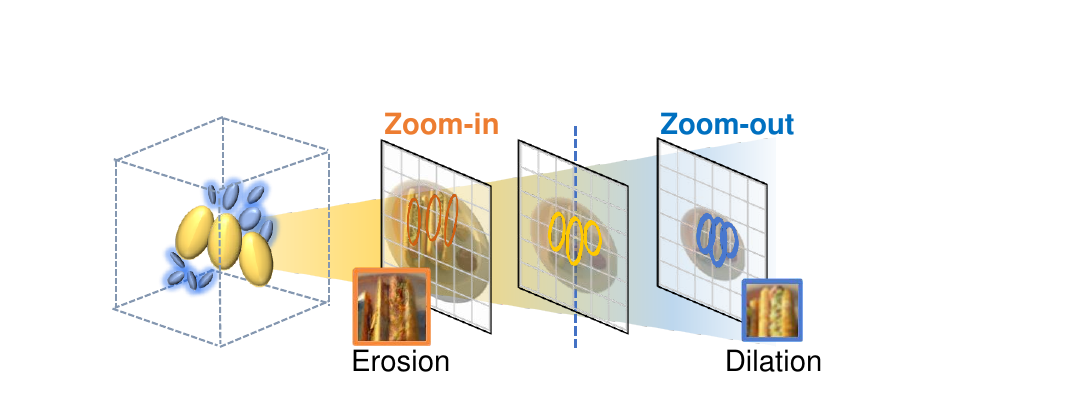}
    \vspace{-1cm}
    \caption{Aliasing at different observation distances.}
    \label{fig:issue}
    \vspace{-0.75cm}
\end{figure}

\begin{figure*}[ht]
    \vspace{-1cm}
    \centering
    \includegraphics[trim=1.2cm 5cm 0.8cm 5cm,clip=true,width=0.95\linewidth]{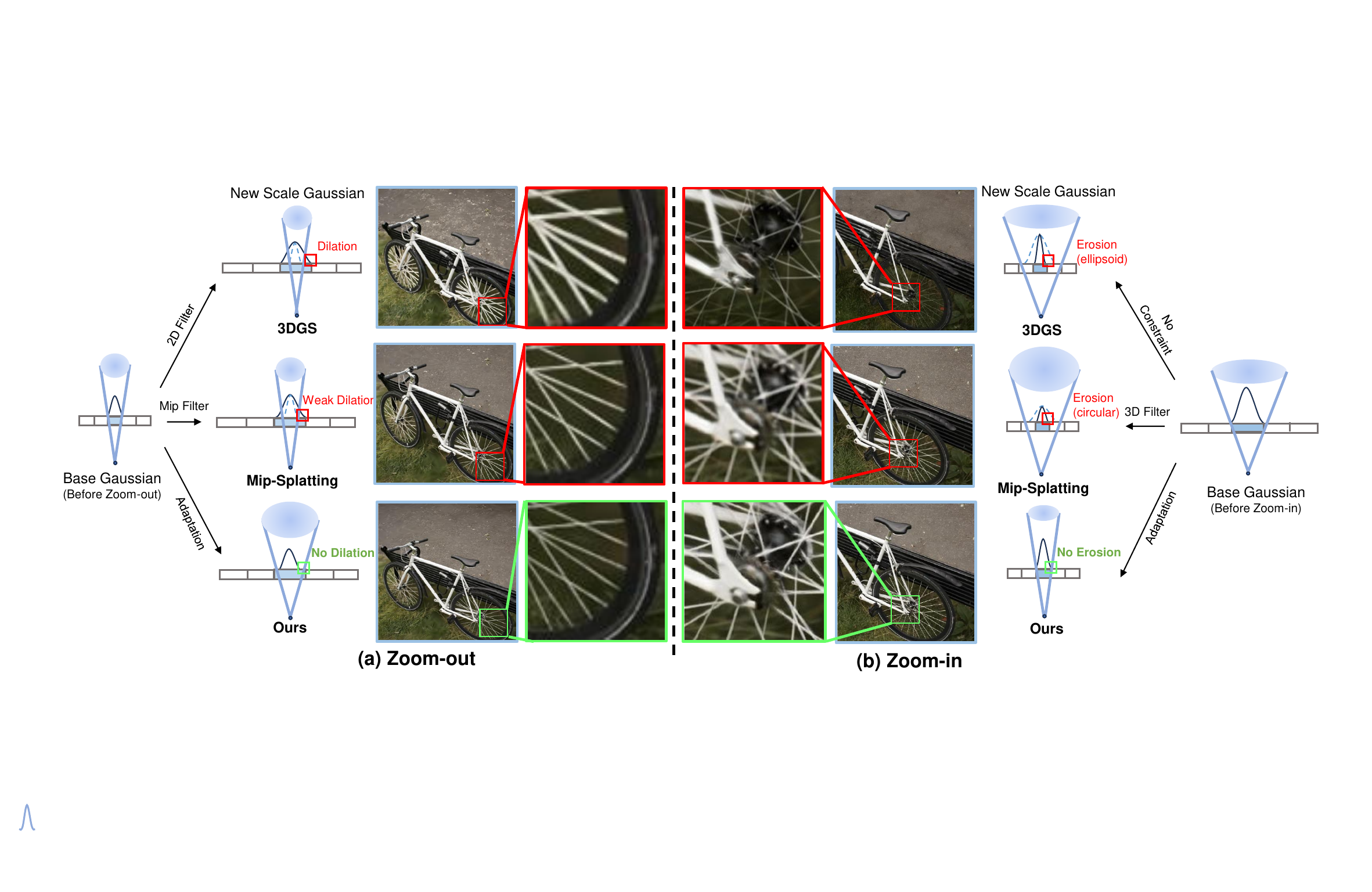}
    \label{fig:fig1_all}
    \vspace{-0.9cm}
    \caption{\textbf{Comparison of 3DGS, Mip-Splatting and ours in zoom-out and zoom-in scenarios.} 3DGS (top) lacks scale sensitivity and exhibits strong degradation. 
    Mip-Splatting (middle) introduces a 2D Mip filter and 3D smooth filter to regulate scales heuristically while leaving 3D Gaussians unaware of the varying footprint in pixel space.
    Ours (bottom) adopts scale-adaptive Gaussians for explicit pixel coverage optimization. The projection in previous pixel coverage is illustrated in the dashed line.}
    \vspace{-0.4cm}
    \label{fig:compar}
\end{figure*}

With the development of implicit methods \cite{barron2021mip,barron2022mip,muller2022instant,xu2022point}, Novel View Synthesis  (NVS) gains significant attention in virtual reality \cite{zhou2016view,fei20243d,chen2024survey}, augmented reality \cite{wang2021ibrnet, shi2021self,riegler2020free}, and 3D generation \cite{liu2023syncdreamer,poole2022dreamfusion,liu2023zero,tang2023dreamgaussian,yi2023gaussiandreamer,chen2023text,feng2024fdgaussian,ren2023dreamgaussian4d,melas2023realfusion,yu2021pixelnerf}.
Recently, 3D Gaussian Splatting \cite{kerbl20233d} shows state-of-the-art rendering quality with fast speed, due to the primitive-based representation and rasterization technique. However, 3DGS suffers from severe aliasing or blurriness when zooming in or out, leading to compromised user experience in interaction applications. 
Typically trained on single-scale images, 3DGS is sensitive to sampling rates that deviate from training set \cite{yu2023mip,yan2023multi,liang2024analytic}. This characteristic bears similarity to Neural Radiance Field (NeRF) \cite{mildenhall2021nerf}, as discussed in Mip-NeRF \cite{barron2021mip}. When changing the observation distance, image resolution or camera focal length, Fig. \ref{fig:issue} illustrates the zoom-out dilation and zoom-in erosion in 3DGS, \textbf{both of which called aliasing} in our paper following \cite{yu2023mip}.
The quality degradation at unseen scales comes from the mismatch between the learned Gaussian fields and out-of-distribution sampling rates. 
Specifically, the 3D Gaussians are projected into the image plane and go through a 2D dilation filter before the rasterizer to smooth the shrinkage bias \cite{yu2023mip}.
However, non-deformable Gaussians and constant dilation factor (which is set to be 0.3) do not seamlessly align with varying settings in Fig. \ref{fig:compar}.
When zooming in, the shrunk Gaussians in screen space are not adequately smoothed, leading to needle-like spikes on the screen. 
Even worse, the finer pixel gridding in zoom-in leads to empty areas where no Gaussian is splatted or shaded, leading to structure missing.
Conversely, zooming out encounters excessive brightness and thickness, since too many Gaussians contribute to a single pixel.

To address the challenges of zooming in and out, most existing methods have unsatisfactory performance due to the lack of deformable representation.
To accommodate the decreased sampling rate in zoom-out, \cite{yan2023multi} proposes a selective rendering method, which aggregates many small primitives for a few large ones locally. 
However, the selection process inevitably introduces manually selected parameters and only applies in zoom-out. 
Inspires by \cite{zwicker2001ewa}, \cite{yu2023mip} introduces a 3D filter for frequency constraint, which works to smooth the zoom-in shrink, and replaces the original 2D dilation filter with a Mip filter inspired by \cite{zwicker2001ewa}. Although applicable for arbitrary scale rendering, the Mip filter in \cite{yu2023mip} lacks scale information and the 3D smooth filter suppresses the high-frequency components, which inevitably sacrifices some finer details \cite{liang2024analytic}.  
\textbf{Thus, there is an urgent need to explore methods that adapt the source signal itself to essentially accommodate varying sampling rates.}

In this paper, we propose a novel scale-adaptive optimization method for 3D Gaussian Splatting, addressing the challenge of zoom-in and zoom-out rendering. 
Specifically, we construct mipmap-like pseudo ground-truth (pseudo-GT) to provide scale-specific information in a self-supervised way. 
A mipmap represents a signal (e.g. image or texture map) at a set of discrete downsampling scales and selects the appropriate scale for anti-aliasing rendering \cite{barron2021mip}. We change the once-created and precomputed mipmap to scale-specific mipmap-like pseudo-GT on novel views during test time.
To deform Gaussians with the mipmap, we introduce a scale-aware guidance loss.
By self-supervised optimization, our method fits for arbitrary-scale rendering.
Our contributions are summarized as follows:
\begin{itemize}[leftmargin=3mm]
    \item We propose a scale-adaptive optimization approach with scale-aware guidance loss. Different from existing methods, our deformable Gaussians are self-adjusted for zoom-in and -out.
    \item Our method designs mipmap-like pseudo-GT on observation scales for test-time adaptation, which provides prior information for zooming deformation. 
    \item Rather than training from scratch, our approach makes use of base Gaussians and converges within 1K iterations, offering computational advantages.
    \item Serving as a plug-in module, our method is applicable in any 3DGS models to address the zoom-in and -out issues, and improves the PSNR of 3DGS over 9dB.
\end{itemize}
\section{Related Work}
\subsection{Novel View Synthesis}
Novel View Synthesis aims to generate images from new viewpoints based on a set of source-view images, enabling the creation of previously uncaptured views and scene reconstruction.
We will introduce two main branches according to different scene representations, \textit{i.e.} explicit and implicit methods.

Explicit methods describe the scene in discrete spatial structures like point cloud \cite{wiles2020synsin,wang2019mvpnet,qi2017pointnet}, mesh \cite{hu2021worldsheet} and voxel grid \cite{liu2020neural}. Yet, explicit methods require large memory usage and perform not well on low-resolution scenes \cite{sun2022direct,fridovich2023k}.

As a pioneer of implicit representations, NeRF \cite{mildenhall2021nerf} utilizes Multi-Layer Perceptron (MLP) to store scene information implicitly, which maps view-related spatial positions and view directions to color and volume density values. 
To speed up the training or rendering of NeRF, \cite{fridovich2022plenoxels,chen2022tensorf,muller2022instant,xu2022point} represent the scene by grid-based feature, point-based feature and hash encoding. 
In multi-scale rendering works \cite{barron2021mip,barron2022mip,nam2024mip,hu2023multiscale,hu2023tri}, Mip-NeRF \cite{barron2021mip} mitigates aliasing at smaller rendering resolutions by replacing pixel-wise ray sampling and rendering of NeRF with conical frustums. 
Despite the significant performance of NeRFs, the volumetric ray marching necessitates high computation afford. 

\subsection{3D Gaussian Splatting}
 Combining explicit and implicit methods for scene representation \cite{chen2024survey},  3DGS \cite{kerbl20233d} learns a Gaussian field that holds discrete geometry centers and continuously optimized attributes. 
Due to its supervisor fidelity and speed, Gaussian Splatting \cite{kerbl20233d} appears to be an appealing alternative to NeRF \cite{barron2021mip}.
 Unlike the MLP architecture and ray marching techniques in NeRF, 3DGS employs primitive-based splatting and fast rasterizer, enjoying high parallelism in GPU \cite{yifan2019patch,kerbl20233d}. The learnable parameters in 3D Gaussians are directly optimized, including geometry information of mean and covariance matrix, and color information including opacity and spherical harmonic (SH) coefficients. The growth and pruning of Gaussians are conducted by adaptive density control considering the accumulated gradients and opacity contributions, which adjust the scene coverage and granularity efficiently.

3DGS has been applied in many tasks, such as sparse-view reconstruction \cite{chung2023depth, yang2024gaussianobject,charatan2023pixelsplat,chen2024mvsplat,wewer2024latentsplat}, 3D generation \cite{xu2024agg,tang2024lgm,zhang2024gaussiancube} and dynamic scenes \cite{yang2023deformable,bae2024per,qian20243dgs,guo2024motion}.
To improve the rendering quality, Pixel-GS \cite{zhang2024pixel} proposes a scaled gradient field optimized by pixel coverage weighted loss, while \cite{zhang2024fregs} introduces frequency spectrum as guidance.
\cite{hamdi2024ges} put forward generalized exponential splatting to sharpen high-frequency details, and \cite{huang20242d} simplifies 3D Gaussians to 2D Gaussians for surface reconstruction. 
Scaffold-GS \cite{lu2023scaffold} leverages anchor points and MLP to arrange view-adaptive Gaussians, which also brings storage benefits.
\cite{ye2024absgs,bulo2024revising,fan2024trim} revise the density control strategy of 3DGS during optimization.
\cite{yu2024gaussian} propose Gaussian-Opacity-Field (GOF) for surface alignment and mesh extraction, and a novel densification method.
\cite{wu2024dual} leverages 3DGS to synthesize data factory under different focal lengths.
\subsection{Zoom-out and Zoom-in}
Despite the superior NVS capacity of NeRF and 3DGS, 
they face degradation when rendering at unseen scales, \textit{i.e.}
 zoom-out and zoom-in.

When zooming out, the disparity between decreased sampling rate and high-frequency components causes aliasing. Mipmap and Level-of-Detail (LOD) techniques are leveraged in traditional computer graphics for anti-aliasing rendering. 
In NeRF-related works, \cite{barron2021mip,barron2022mip,barron2023zip} render conical frustums instead of rays and apply pre-filtering to the input positional encoding to resist aliasing. 
For 3DGS, \cite{ren2024octree,yan2023multi} adopt LoD representation for efficient and anti-aliasing rendering, while inevitably introducing hyperparameter tuning in level selection. 
Analytic-Splatting \cite{liang2024analytic} treats each pixel as an area instead of separate points 
to deal with pixel footprint changes.

Zoom-in entails moving from a global view to local details. 
Previous NeRF-based methods leverage super-sampling strategy with depth regularization \cite{wang2022nerf}, and post-processing steps \cite{lin2024fastsr, huang2023refsr} to address this issue. 
\cite{yu2024gaussiansr,shen2024supergaussian} takes SR prior from images and video respectively for HR rendering.
Mip-Splatting \cite{yu2023mip} introduces a 3D smooth filter to regulate high-frequency components of each primitive. 
\cite{yu2023mip} re-calculates and updates the scale of the 3D filter during training. In the testing stage, each primitive is subjected to the integrated 3D filter before being projected to screen space. 
Unlike previous methods, we propose scale-specific supervision to mimic mipmap. Rather than storing a precomputed data structure and interpolating between levels,  we derive mipmap-like pseudo-GT on the fly and use it to refine the 3D model further.
\section{Preliminaries}
\subsection{3D Gaussian Splatting}
\label{subsec:preliminary-1}

3DGS \cite{zwicker2001ewa,kerbl20233d} represents scenes as a set of 3D Gaussian primitives $\{G_k \,|\, k = 1, \ldots, K\}$, with geometry center $\mu_k \in \mathbb{R}^{3 \times 1}$ and covariance matrix $\Sigma_{k} \in \mathbb{R}^{3 \times 3}$. Each 3D Gaussian is parameterized as:
\begin{equation}
G_{k}(x) = e^{-\frac{1}{2}(x-\mu_{k})^{\top} \Sigma_{k}^{-1} (x-\mu_{k})},
\end{equation}
where $\Sigma_{k}$ is further expressed by orthogonal rotation matrix $R \in \mathbb{R}^{3 \times 3}
$ and diagonal scale matrix $S \in \mathbb{R}^{3 \times 3}$:
$\Sigma = RSS^{\top}R^{\top}$.
During rendering, primitives overlapped at pixel $x$ are sorted according to depth order $\{1, \ldots, K\}$. 
Then the view-dependent pixel color $c(x)$ modeled by spherical harmonics is calculated by :
\begin{equation}
c(x) = \sum_{k=1}^{K} c_k\alpha_k \prod_{i=1}^{k-1}(1-\alpha_i),
\label{eq:shading}
\end{equation}
where $\alpha_k$ is calculated by evaluating a projected 2D Gaussian $G_k^{2D}$ multiplied with a learned opacity \cite{yifan2019patch}, and finally attends to the pixel color $c(x)$ with the primitive color $c_k$.
Each primitive properties are optimized across all the training views coupled with density control intermittently.
\vspace{1mm}
\noindent\textbf{Shrinkage bias and 2D dilation.}
In practice, 3D Gaussians are truncated to calculate pixel color with shrinkage bias \cite{yu2023mip}. The extreme shrink is expressed as $\delta$ impulse, which becomes needle-like spikes in renderings. In addition, the tiny Gaussians (smaller than 1 pixel) with trivial contributions are hard to optimize. Therefore, a 2D dilation operation is designed before rasterization to undo the shrinkage bias:
\begin{equation}
G^{2D}_k(x) = e^{-\frac{1}{2}(x-\mu_k)^{\top}(\Sigma^{2D}_k + sI)^{-1}(x-\mu_k)},
\end{equation}
where $s$ is set to be 0.3. Working as a low-pass filter, the dilation operation smoothes the shrinkage bias to achieve faithful rendering in Fig. \textcolor{red}{3a}.

\begin{figure}[!ht]
    \centering
    \vspace{-0.3cm}
        \includegraphics[trim=5.5cm 7cm 5.5cm 7.2cm,clip=true,width=\linewidth]{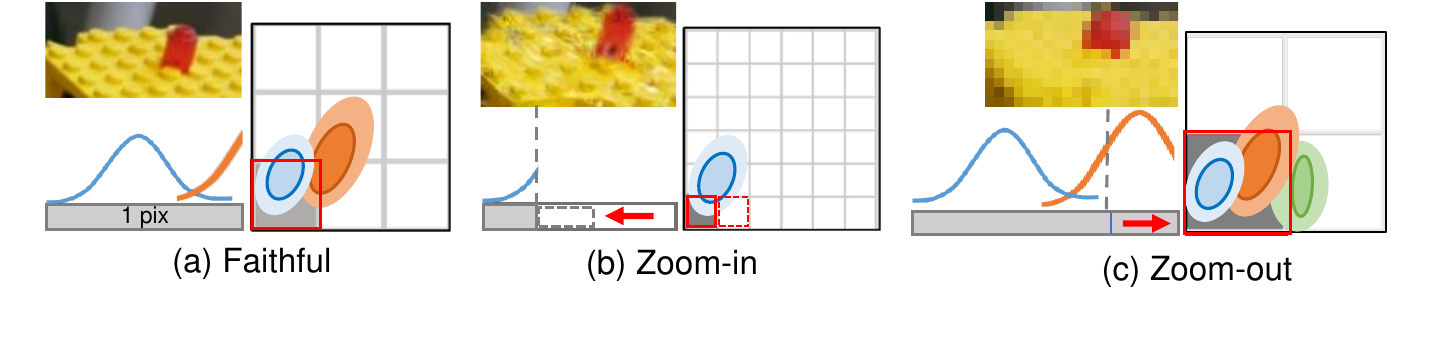}
    \vspace{-1cm}
    \caption{\textbf{A toy 3DGS model under multi-scale rendering.} The splatted Gaussians (dark) go through 2D dilation (light) for faithful rendering (a). However, the varying shrinkage bias and constant dilation cause zoom-in spikes (c) and zoom-out thickness (d).
    }
    \vspace{-0.3cm}
    \label{fig:signal}
\end{figure}
\subsection{Motivation}
\label{subsec:preliminary-2}

\begin{figure*}[t]
    \centering
    \vspace{-0.9cm}
    \includegraphics[trim=5.65cm 4cm 4.5cm 3cm,clip=true,width=\linewidth]{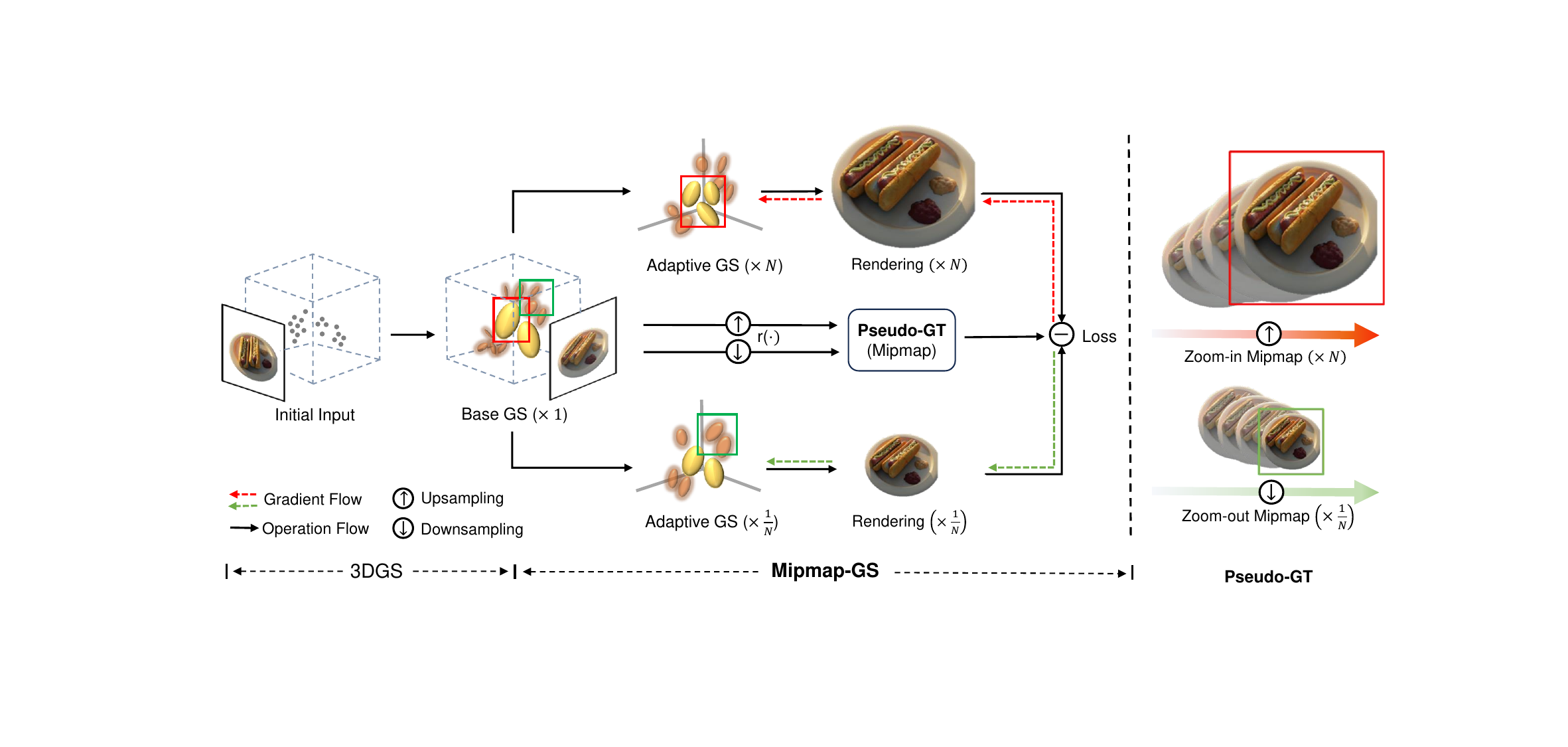}
        \vspace{-0.7cm}
        \caption{\textbf{Pipeline}. We adjust single-scale trained base Gaussians for arbitrary zoom factors supervised by mipmap-like pseudo-GT. The core component is the mipmap proposal module, which constructs scale-specific mipmap by upsampling or downsampling novel views. }
        \label{fig:pipeline}
        \vspace{-0.4cm}
\end{figure*}
Despite the promising NVS results at basic scale (consistent with the training scale), zoom-in and -out rendering will bring two challenges for out-of-distribution generalization:

\vspace{1mm}
\noindent\textbf{Varying sampling rates. }
3DGS leverages a group of Gaussian ellipsoids for scene representation, akin to a Gaussian mixture model.
During pixel shading, pixel grids are allocated according to the current sampling rate. Then, Gaussians overlapped in one pixel are accumulated with shrinkage bias \cite{yu2023mip}. When the sampling rate changes, the varying shrinkage bias leads to confusion during accumulation.
In Fig. \ref{fig:signal}, we simply illustrate different sampling rates and the shrunk Gaussian signals in 1D. The learned Gaussian field samples two signals at basic scale (Fig. \textcolor{red}{3a}), while a zoom-in pixel (Fig. \textcolor{red}{3b}) only captures a blue signal, and a zoom-out pixel (Fig. \textcolor{red}{3c}) accumulates the orange signal for excessive contribution.

\vspace{1mm}
\noindent\textbf{Constant dilation. }
The pixel coverage during shading is decided jointly by projected Gaussians and dilation operation.
The 2D dilation operation in 3DGS \cite{kerbl20233d} is designed to amplify tiny Gaussians (marked as blue) smaller than one pixel (marked as grey) in Fig. \textcolor{red}{3a}.
However, the constant dilation doesn't fit for decreased pixel coverage in zoom-in, causing thin structures and empty areas with no splats (Fig. \textcolor{red}{3b}). Conversely, maintaining the same dilation for zoom-out leads to excessively wide coverage and thick structures, which also undermines the rendering efficiency. In Fig. \textcolor{red}{3c}, the green Gaussian was not meant to fall into the grey pixel, but due to inherited dilation, it gets included for pixel shading. The intuitive solution of super-sampling is sub-optimal due to the computational burden.

To mitigate the misalignment, 
\cite{yu2023mip} conducts both 3D and 2D smoothing using two smoothing filters, while \cite{song2024sa} introduces a scale-adaptive 2D filter.
These filters partially revise the scale of Gaussians to fit multi-scale rendering, yet the overall distribution doesn’t actually change.
\cite{yan2023multi,ren2024octree} arrange Gaussians hierarchically by LoD architecture, while the level selection is not accurate for pixel coverage and performs sensitive to hyper-parameters.  
Therefore, it's impossible to fit multi-scale rendering with fixed Gaussians. We propose Mipmap-GS to provide deformable Gaussians optimized via scale-specific mipmap.
\vspace{-0.2cm}
\section{Proposed Method}
To solve the mismatched pixel coverage and Gaussians at new scales, we deform Gaussians interactively to fit for varying zoom factors. First, we construct mipmap-like pseudo-GT at the observation scale, see Sec. \ref{subsec:Mipmap-like Pseudo-GT}.  Then, we introduce scale-adaptive Gaussians in Sec. \ref{subsec:Scale-Adaptive 3D Gaussians}, which are deformed from base Gaussians. Finally, we describe optimization details in Sec. \ref{subsec:Optimization}, including scale-aware guidance loss and active pruning strategy.
\subsection{Mipmap-like Pseudo-GT}
\label{subsec:Mipmap-like Pseudo-GT}
Instead of conducting time-consuming multi-scale training, our method optimizes scale-adaptive Gaussians using mipmap-like pseudo-GT.
Considering the NVS capacity of 3DGS, we first splat the set of base Gaussians into novel views at the basic scale (${\times}1$), which faces no degradation. 
Then, we construct new scale pseudo-GT with mipmap resizing function $r(x)$.
For zoom-in adaptation, novel-view renderings are upsampled to ${\times}N$ using super-resolution methods like SwinIR \cite{liang2021swinir}. 
Similarly, to create a scale-specific mipmap for zoom-out, the rendered images are downsampled to ${\times}1/N$ of the original resolution. Compared with the zoom-in process which needs to generate details for extended pixels, the downsampling step involves less textural deterioration to mimic fading away. Therefore, we simply adopt bilinear interpolation as $r(x)$ to generate the LR mipmap.
Note that mipmap is originally defined as a pre-calculated image sequence with progressively lower resolutions. 
Here, we use mipmap to refer to pseudo-GT at both lower and higher resolutions.


\subsection{Scale-Adaptive 3D Gaussians}
\label{subsec:Scale-Adaptive 3D Gaussians}
\begin{figure*}[t]
    \centering
    \vspace{-0.9cm}
    \includegraphics[trim=1.7cm 1cm 2cm 3.6cm,clip=true,width=0.9\linewidth]
    {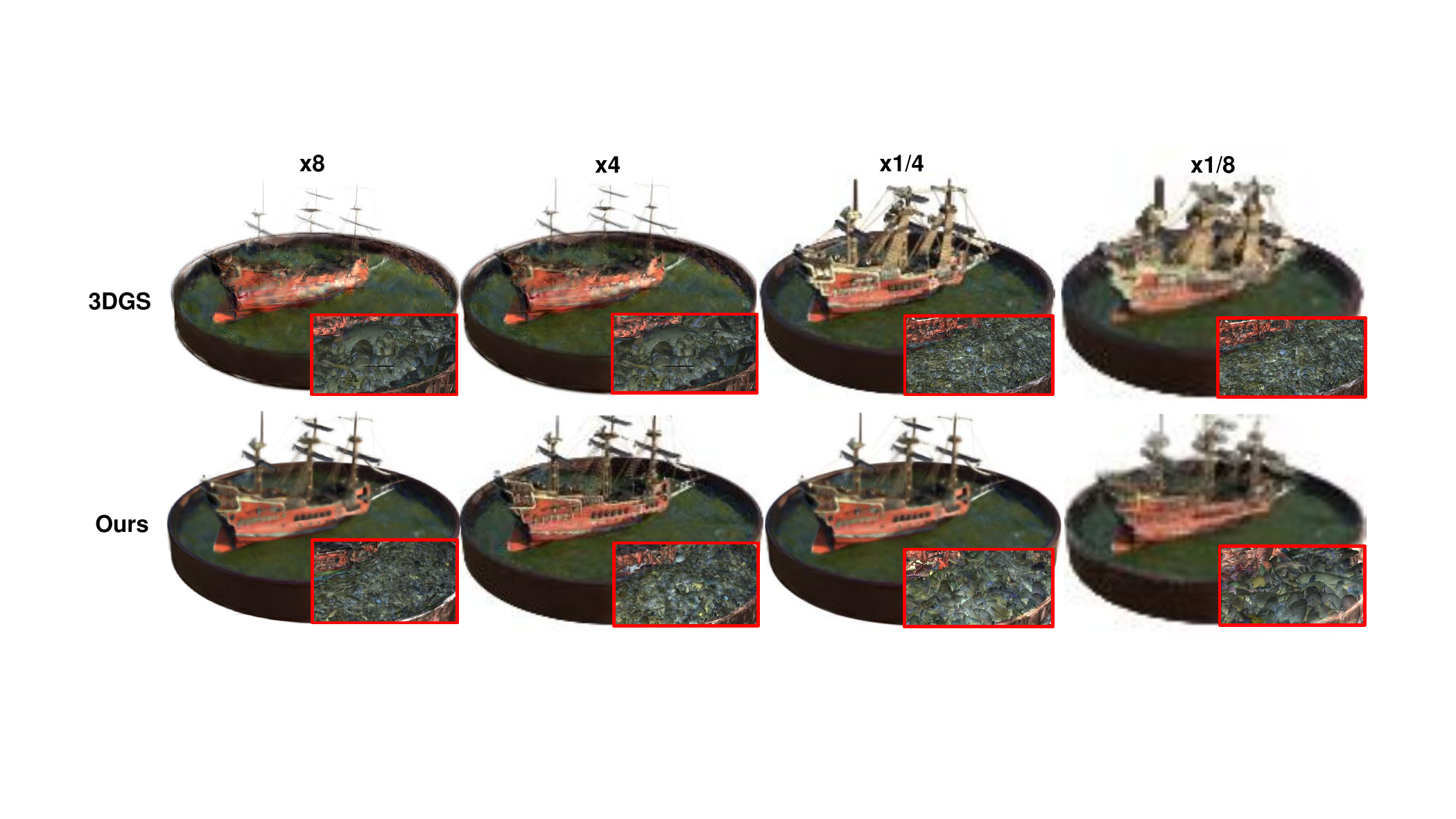}
    \vspace{-2cm}
    \caption{\textbf{Visualization of Gaussians.} 3DGS \cite{kerbl20233d} use fixed Gaussians for multi-scale rendering, while our adaptive adjustment provides deformable Gaussians towards various zoom factors.}
    \vspace{-0.5cm}
    \label{fig:visualiz}
\end{figure*}
As introduced in Sec. \ref{subsec:preliminary-1}, 
3DGS \cite{kerbl20233d} projects Gaussian ellipsoids into image space as a set of $\alpha$-blended 2D Gaussians to accumulate the pixel color. The 3D shrinkage bias and 2D dilation restoration cooperate to fit the training scale. 
However, taking place in different spaces, these two operations work independently while not seamlessly aligned.
As illustrated in Sec. \ref{subsec:preliminary-2}, the existing space gap causes two challenges at multi-scale rendering,
since dramatically varied sampling rates exceed the control range of the constant dilation, leading to aliasing results in Fig. \ref{fig:visualiz}.
To bridge the gap, we create the scale-specific mipmap $r(\hat{x})$ from base Gaussians $\mathbf{G}$.
Then, we deform $\mathbf{G}$ into optimized Gaussians $\mathbf{G}^{\text{opt}}$ by photometric loss between the rendered image $x$ and scale-specific mipmap $r(\hat{x})$: 
\begin{equation}
 \mathbf{G}^{\text{opt}} = \mathbf{G} - \beta
 \nabla L(x, r(\hat{x})),
\end{equation}
where $\beta$ is the learning rate.
Notably, our method solves the aliasing essentially since the scale prior directly informs of the new sampling rate to address two challenges in Sec. \ref{subsec:preliminary-2}, instead of introducing filters to ease the issues caused by constant dilation.
The scale-adaptive Gaussians leverage deformable attributes and distribution to fit any scale rendering.
As a test-time adaptation method, our method keeps the scale consistency between 3D primitives and 2D pixel coverage for wide-coverage inference contexts.

\subsection{Optimization}
\label{subsec:Optimization}
Illustrated in Alg. \ref{algo:adaptive_optimization}, we first generate the novel view $\hat{x}$ by base Gaussians $\mathbf{G}$. Then, $\hat{x}$ is upsampled or downsampled into $r(\hat{x})$ via mipmap function, serving as the pseudo-GT.
We define the scale-aware guidance loss in adaptation as: 
\begin{equation}
L(x, r(\hat{x})) = ||x-r(\hat{x})||^2.
\end{equation}
Here we use $\ell_2$ loss and discuss other losses in Sec. \ref{subsec: ablation_study}. The deformation from $\mathbf{G}$ to ${\mathbf{G}^{\text{opt}}}$ is finished within 1K iterations.


    
    
     



Except for performance improvement, we also realize compact scene representation. In Alg. \ref{algo:adaptive_optimization}, the optimized primitive number $K^{\text{opt}}$ of ${\mathbf{G}^{\text{opt}}}$ is smaller than the previous number $K$.
The density control in 3DGS \cite{kerbl20233d} stops halfway through the optimization process, leaving many low opacity Gaussians with trivial contribution \cite{lu2023scaffold,fan2023lightgaussian,lee2023compact,morgenstern2023compact}. 
In contrast, we keep active pruning during the whole process for a more compact representation, see more in supplementary materials. 
The storage size reduction by our active pruning is more obvious in zoom-out situations where the decreased spatial resolution requires fewer Gaussians to represent. 
What's more, our adjustment needs merely $ 3\% $ iterations compared with training from scratch.
Bringing neither interference towards the original pipeline nor new hyper-parameters, our method is compatible with subsequent works like Scaffold-GS \cite{lu2023scaffold}, shown in Tab. \ref{tab:zoom-inout-blender-l1-ssim}.


\vspace{-0.4cm}
\begin{algorithm}[h!]
\caption{Adaptive optimization of our method}
\label{algo:adaptive_optimization}
\KwIn{base Gaussians $\{\mathbf{G}_k \,|\, k = 1, \ldots, K\}$, viewpoints $\{V_j \,|\, j = 1, \ldots, J\}$, scale $N$, iteration $S$}
\KwOut{optimized Gaussians $\{\mathbf{G}^{\text{opt}}_k \,|\, k = 1, \ldots, K^{\text{opt}}\}$ }
\SetAlgoLined
\SetKwInOut{Parameter}{Parameters}
\For{$i=0,...,S$}{
    select $v_j \in {V_j}$ randomly 

    render $\{\mathbf{G}_k\}$ to basic scale image $ \hat{x}^j$, see Eqn. \eqref{eq:shading}
    
    construct mipmap $r(\hat{x}^j)$ by ${\times}N$ upsampling (zoom-in) or ${\times}\frac{1}{N}$ downsampling (zoom-out)
    
    render $\{\mathbf{G}^{i}_k\}$ to new scale image $x^j$, see Eqn. \eqref{eq:shading}
     
    optimize $\{\mathbf{G}^{i}_k\} = \mathop{\arg\min}\nolimits_{\substack\{\mathbf{G}^{i}_k\}} L(x^j, r(\hat{x}^j))$
    
    \If{$i \bmod 100 == 0$}{
    densify or prune
    }
}

\Return{$\{\mathbf{G}^{\text{opt}}_k\} = \{\mathbf{G}^{S+1}_k\}$}
\end{algorithm}
\vspace{-0.7cm}
\section{Experiments}
\label{sec:exp}

\begin{table*}
\begin{subtable}{1\linewidth}
\vspace{-0.7cm}
    \centering
        \begin{tabularx}{\linewidth}{l*{9}{X}}
        \hline
        \multirow{2}{*}{Method} & \multicolumn{3}{c}{\textbf{${\times}1/2$}} & \multicolumn{3}{c}{\textbf{${\times}1/4$}} & \multicolumn{3}{c}{\textbf{${\times}1/8$}} \\ 
        & SSIM & PSNR & LPIPS & SSIM & PSNR & LPIPS & SSIM & PSNR & LPIPS  \\ \hline
        NeRF \cite{mildenhall2021nerf} & 0.962   & 32.43 & 0.041 & 0.964 & 30.29 &	0.044  & 0.951 & 26.70& 0.067  \\
        Mip-NeRF \cite{barron2021mip}& 0.970 & 33.31 & 0.031 & 0.969 & 30.91 &	0.036 &  0.961 & 27.97 & 0.052 \\
        Instant-NGP \cite{muller2022instant}& 0.969  & 33.00 & 0.033 & 0.964 & 29.84 &	0.046 &  0.947 & 26.33 & 0.075 \\    \hline    
        3DGS \cite{kerbl20233d} & 0.951 & 27.14 & 0.031 & 0.875 &	21.39	& 0.067 & 0.763 &	17.59 &	0.127 \\
        3DGS+EWA \cite{kerbl20233d,zwicker2001ewa}& \cellcolor{orange}0.971 & 31.66 & 0.024 & 0.959 & 27.82 & 0.033 & 0.940 & 24.62 & 0.047 \\
        Scaffold-GS \cite{lu2023scaffold}& 0.953 & 27.48 & 0.030 & 0.886 & 21.83 & 0.061 & 0.781 & 18.03 & 0.116\\
        Pixel-GS \cite{zhang2024pixel}& 0.947&27.96&0.032&0.866&	22.67&	0.070&	0.748&	19.09&	0.133\\
        Octree-GS \cite{ren2024octree} &0.950&27.30&0.061&0.888&21.97&0.063& 0.787&	18.15	&0.114\\
        Analytic-Splatting \cite{liang2024analytic} &\cellcolor{peach}0.977& \cellcolor{peach}34.21 &\cellcolor{peach}0.019 &\cellcolor{orange}0.977& 31.49& \cellcolor{orange}0.021 &0.969& 28.42& 0.031\\
        Mip-Splatting \cite{yu2023mip} $\qquad\qquad$ & \cellcolor{peach}0.977 & 34.00 & \cellcolor{peach}0.019 & \cellcolor{peach}0.978&31.85&\cellcolor{peach}0.019	&\cellcolor{orange}0.973 &28.67 & \cellcolor{peach}0.026\\
       Scaffold-Ours &\cellcolor{orange}0.971 & 32.71 & \cellcolor{orange}0.026 &0.975 &\cellcolor{orange}32.46 & 0.023 &  \cellcolor{peach}0.976 &\cellcolor{peach}30.81 &\cellcolor{orange}0.028\\ 
      3DGS-Ours & \cellcolor{peach}0.977 &	\cellcolor{orange}34.18 & \cellcolor{peach}0.019 & \cellcolor{peach}0.978 & \cellcolor{peach}32.51 & \cellcolor{orange}0.021 & \cellcolor{peach}0.976 &	\cellcolor{orange}30.64 & \cellcolor{peach}0.026 \\  \hline
    \end{tabularx}
    \vspace{0.025cm}
    \caption{Zoom-out.}
    \label{tab:zoom-out-blender}

\end{subtable}

\begin{subtable}{1\linewidth}
    \centering
        \begin{tabularx}{\linewidth}{l*{9}{X}}
        \hline
        \multirow{2}{*}{Method} & \multicolumn{3}{c}{\textbf{${\times}2$}} & \multicolumn{3}{c}{\textbf{${\times}4$}} & \multicolumn{3}{c}{\textbf{${\times}8$}}\\ 
        & SSIM & PSNR & LPIPS & SSIM & PSNR & LPIPS & SSIM & PSNR & LPIPS \\ \hline
        NeRF \cite{mildenhall2021nerf}& 0.921 & 27.54 & 0.100& 0.881& 25.56 &0.170 & - & -& -  \\
        Mip-NeRF \cite{barron2021mip} & 0.944&29.36&0.057&0.876&25.47&0.159&0.832&23.47&0.207\\
        NeRF-SR \cite{wang2022nerf}& 0.946 & 29.77 &0.045 & 0.921 &28.07 &0.071 & - & -& - \\ \hline  
        3DGS \cite{kerbl20233d}& 0.907 &	23.38 &	0.068  &  0.832 &	19.93&0.128&  0.824	& 18.52	&0.153  \\
        Scaffold-GS \cite{lu2023scaffold} & 0.770 & 21.14 & 0.097 & 0.800 & 17.35 & 0.163 & 0.807 & 16.01 & 0.177\\
        Pixel-GS \cite{zhang2024pixel}&0.897&24.95&0.070&0.823&21.26&	0.133& 0.819&	19.69&	0.161\\
        Mip-Splatting \cite{yu2023mip}& 0.960 & 30.08& 0.051  & 0.917 &27.12& 0.105 &0.886 & 25.71&\cellcolor{orange}0.136 \\
        Scaffold-Ours & \cellcolor{peach}0.968 &\cellcolor{orange}30.79 &\cellcolor{orange}0.043 &\cellcolor{peach}0.936	&\cellcolor{orange}28.20 &\cellcolor{peach}0.078 &\cellcolor{peach}0.911 &\cellcolor{peach}26.59 &0.153\\
         3DGS-Ours& \cellcolor{orange}0.964 & \cellcolor{peach}31.23 & \cellcolor{peach}0.041 & \cellcolor{orange}0.927 & \cellcolor{peach}28.29 & \cellcolor{orange}0.081 & \cellcolor{orange}0.897 & \cellcolor{orange}26.43 & \cellcolor{peach}0.119\\
        \hline
    \end{tabularx}
    \vspace{0.025cm}
    \caption{Zoom-in.}
    \label{tab:zoom-in-blender}
    \vspace{-0.4cm}
\end{subtable}
    \caption{\textbf{Zoom-out and zoom-in comparisons on NeRF Synthetic dataset} \cite{barron2021mip}. \textcolor{peach}{JJJJ}}
    \label{tab:zoom-inout-blender-l1-ssim}
    \vspace{-0.5cm}
\end{table*}

\subsection{Experimental Setup} 
\vspace{1mm}
\noindent\textbf{Datasets.}
We conduct experiments on the NeRF Synthetic dataset \cite{mildenhall2021nerf} and the Mip-NeRF 360 dataset \cite{barron2021mip}, measuring the effect of our continuous optimization approach. The Blender dataset has 8 synthetic objects with no background, which is widely used in NeRF methods. And we adopt the common split of 100/100/200 for training/validation/test. 
The more challenging Mip-NeRF 360 dataset \cite{barron2022mip} consists of 9 real-world unbounded scenes. We follow the split factor 8 and report the average metrics over all scenes.

\vspace{1mm}
\noindent\textbf{Evaluation metrics.}
The rendering quality at each viewpoint is evaluated by Structural Similarity Index Measure (SSIM), Peak Signal-to-Noise Ratio (PSNR) and LPIPS \cite{zhang2018unreasonable}. 
Higher SSIM and PSNR values and lower LPIPS scores indicate better-rendering results.

\vspace{1mm}
\noindent\textbf{Baselines.}
We take 3DGS \cite{kerbl20233d}, Scaffold-GS \cite{lu2023scaffold} and Pixel-GS \cite{zhang2024pixel} as the baseline Gaussian Splatting models. 
In addition, Mip-Splatting \cite{yu2023mip}, Analytic-Splatting \cite{liang2024analytic} and Octree-GS \cite{ren2024octree} are treated as SOTA anti-aliasing methods based on 3DGS. 
We also test the performance of GOF \cite{yu2024gaussian}, since it inherits the anti-aliasing functions from \cite{yu2023mip}. 3DGS + EWA \cite{zwicker2001ewa} replaces the original dilation with the EWA filter \cite{zwicker2001ewa}, introduced in \cite{yu2023mip}.
Following the setting in their paper, \cite{kerbl20233d, lu2023scaffold,zhang2024pixel,yu2023mip,liang2024analytic,yu2024gaussian} are trained for 30K iterations, and \cite{ren2024octree} for 40K iterations. 

\vspace{1mm}
\noindent\textbf{Implementation Details.}
All the experiments are conducted on a single A100 GPU. 
The settings of density control, learning rate schedule and hyper-parameters are consistent with the original paper. After single-scale training, we regard rendering the set of Gaussians directly to different resolutions as the baseline setting following \cite{yu2023mip}. We also test Scaffold-GS \cite{lu2023scaffold} with our adaptation to illustrate the plugin effects of our method.
 \begin{figure}[h!]
    \centering
    \vspace{-0.4cm}
        \includegraphics[trim=7.55cm 6.7cm 7cm 5.9cm,clip=true,width=1\linewidth]{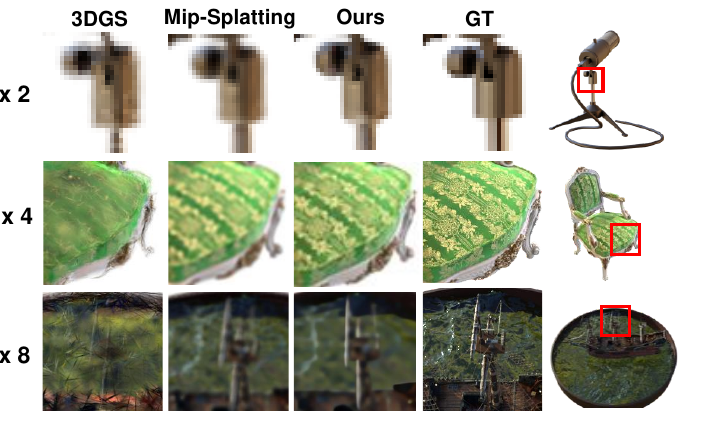}
        \vspace{-1cm}
        \caption{\textbf{Comparison of zoom-in results on NeRF Synthetic dataset} \cite{barron2021mip}.}
        \label{fig:zoom-in-mic}
        \vspace{-0.5cm}
\end{figure}
\subsection{Comparisons with State-of-the-Art}
\begin{table*}
\vspace{-0.7cm}
\begin{subtable}[b]{1\linewidth}
    \centering
        \begin{tabularx}{\linewidth}{l*{9}{X}}
    \hline
    \multirow{2}{*}{Method} & \multicolumn{3}{c}{\textbf{${\times}1/2$}} & \multicolumn{3}{c}{\textbf{${\times}1/4$}} & \multicolumn{3}{c}{\textbf{${\times}1/8$}}\\ 
    & SSIM & PSNR & LPIPS & SSIM & PSNR & LPIPS & SSIM & PSNR & LPIPS \\ \hline
    Mip-NeRF 360 \cite{barron2022mip} & 0.864 & 29.19 & 0.136 &0.912 & 30.45 & 0.077 &  0.931 & 30.86 & 0.058 \\
    Instant-NGP \cite{muller2022instant} & 0.712  & 25.23 & 0.251 &  0.809 & 26.84 &	0.142 &  0.877 & 28.42 & 0.092 \\
    zip-NeRF \cite{barron2023zip}& 0.892   & 30.00 & 0.099 & 0.933 & 31.57 &	0.056  &0.954 & 32.52 & 0.037  \\ \hline
    3DGS \cite{kerbl20233d}&0.774 &26.66 &	0.291 &	 0.721 & 22.87 &	0.241 &	0.763 &	25.36 &	0.250 \\
    Octree-GS \cite{ren2024octree} &0.793&	27.13&0.261 & 0.805&26.44&0.199&0.784&23.92	&0.182 \\
    Analytic-Splatting \cite{liang2024analytic} &0.713& 25.70&0.350 &0.791& 27.04& 0.216 &\cellcolor{peach}0.880& \cellcolor{peach}29.44& \cellcolor{peach}0.106\\
    Mip-Splatting \cite{yu2023mip}& \cellcolor{orange}0.812 & \cellcolor{orange}27.44 & \cellcolor{peach}0.233
    &  0.799 &27.16 & 0.209 &0.744 &27.03 &	 0.326  \\
    GOF \cite{yu2024gaussian} &  0.803 &27.17& \cellcolor{orange}0.251&\cellcolor{orange}0.825&\cellcolor{orange}27.61&\cellcolor{orange}0.185&0.838&27.21&0.155\\
   3DGS-Ours & \cellcolor{peach}0.854 & \cellcolor{peach}28.57 &	\cellcolor{peach}0.156 & \cellcolor{peach}0.828 &	\cellcolor{peach}28.34 &	\cellcolor{peach}0.199  & \cellcolor{orange}0.854 & \cellcolor{orange}28.57 &	\cellcolor{orange}0.154  \\
    \hline
\end{tabularx}
\vspace{0.025cm}
\caption{Zoom-out.}
\label{tab:zoom-out-colmap}
\end{subtable}
\begin{subtable}[t]{1\linewidth}
    \centering
        \begin{tabularx}{\linewidth}{l*{9}{X}}
    \hline
    \multirow{2}{*}{Method} & \multicolumn{3}{c}{\textbf{${\times}2$}} & \multicolumn{3}{c}{\textbf{${\times}4$}} & \multicolumn{3}{c}{\textbf{${\times}8$}}\\ 
    & SSIM & PSNR & LPIPS & SSIM & PSNR & LPIPS & SSIM & PSNR & LPIPS \\ \hline
    Mip-NeRF 360 \cite{barron2022mip}& 0.727 & 25.18 & 0.260 & 0.670 & 24.16 &	0.370 & 0.706  & 24.10 & 0.428 \\
    Instant-NGP \cite{muller2022instant}&  0.639 & 24.76 & 0.367 & 0.626 & 24.27 & 0.445 & 0.698  & 24.27 &  0.475\\
    zip-NeRF \cite{barron2023zip}& 0.696 & 23.27 & 0.257 & 0.565 & 20.87 &	0.421 & 0.559 & 20.27 & 0.494 \\\hline      
    3DGS \cite{kerbl20233d}& 0.740 & 23.49 & 0.243 & 0.619 & 20.69 & 0.394 & 0.603 & 19.21 & 0.477 \\
    3DGS+EWA \cite{kerbl20233d,zwicker2001ewa}& 0.775  & 25.90 & 0.236 & 0.667 & 23.70 & 0.369 & 0.643 & 22.81 & 0.449 \\
    Mip-Splatting \cite{yu2023mip} &  \cellcolor{peach}0.808 &	\cellcolor{peach}27.39&	\cellcolor{peach}0.205 &  \cellcolor{peach}0.754 & \cellcolor{peach}26.47 & \cellcolor{orange}0.305 & \cellcolor{peach}0.765 &	\cellcolor{peach}26.22 & \cellcolor{peach}0.392 \\
    GOF \cite{yu2024gaussian} & \cellcolor{orange}0.802 & 	27.05& 	\cellcolor{orange}0.206& \cellcolor{orange}0.744& 26.17& 	0.313& 0.758 &	25.90 & 0.411 \\
    3DGS-Ours &  0.799 & \cellcolor{orange}27.17 & 0.209 & \cellcolor{orange}0.744 & \cellcolor{orange}26.22 & \cellcolor{peach}0.302	 & \cellcolor{orange}0.761 & \cellcolor{orange}25.91 & \cellcolor{orange}0.399 \\
    \hline
    \end{tabularx}
    \vspace{0.025cm}
    \caption{Zoom-in.}
    \label{tab:zoom-in-colmap}
    \end{subtable}
\vspace{-0.4cm}
    \caption{\textbf{Zoom-out and zoom-in comparisons on Mip-NeRF 360 dataset} \cite{barron2022mip}. }
    \label{tab:zoom-inout-colmap}
    \vspace{-0.4cm}
\end{table*}
\begin{figure*}[h!]
    \centering
        \includegraphics[trim=1.9cm 5cm 2cm 3.2cm,clip=true,width=\linewidth]{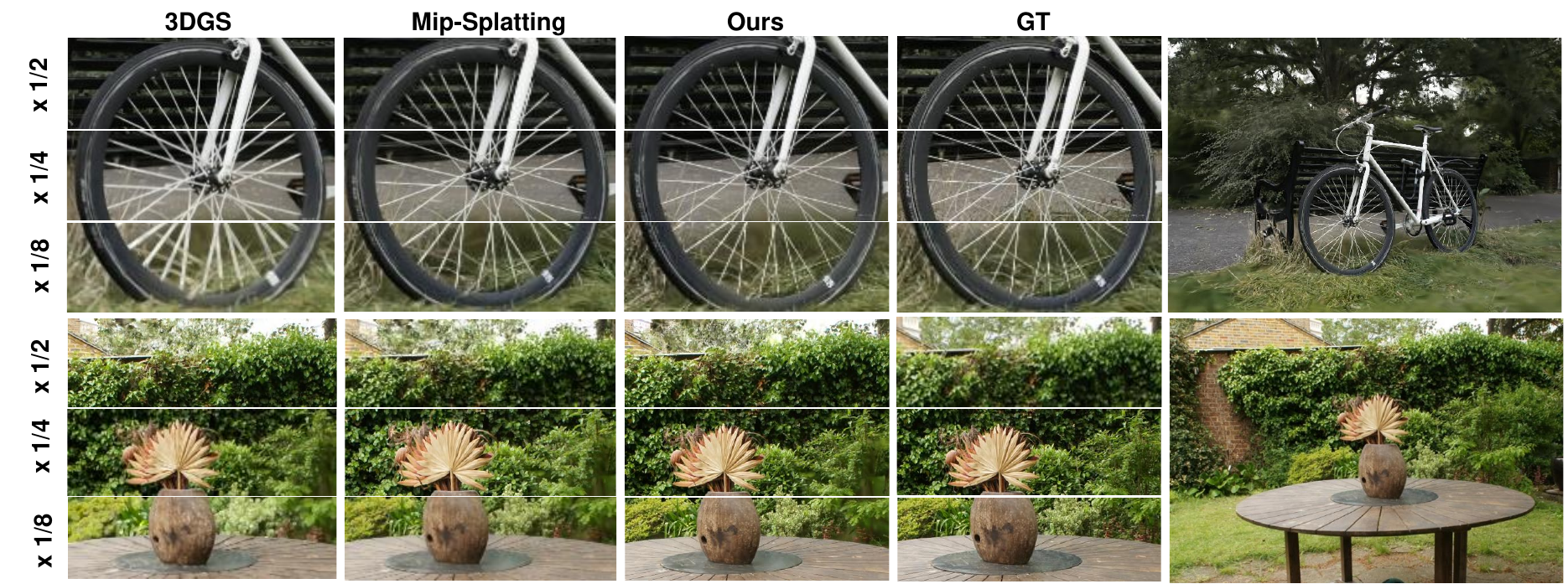}
        \vspace{-0.7cm}
        \caption{\textbf{Comparison of zoom-out results on Mip-NeRF 360 dataset \cite{barron2022mip}.}}
        \label{fig:zoom-out-8687}
        \vspace{-0.5cm}
\end{figure*}
\vspace{-0.2cm}
\noindent\textbf{Zoom-out.}
Following settings introduced by \cite{yu2023mip}, the baseline model is trained on full-resolution (x1) images and rendered at different resolutions (${\times}1/2, {\times}1/4, {\times}1/8$) to simulate the zoom-out effect.
 We derive renderings at the same resolution with respect to training, then downsample them into zoom-out pseudo-GT. Since downsampling from HR images involves minimal loss of information, this process often preserves most of the essential details of the original scene representation. The results in Tab. \ref{tab:zoom-out-blender} and Tab. \ref{tab:zoom-out-colmap} show the improvement after our dynamic optimization. 
 The rendering results at ${\times}1/2, {\times}1/4, {\times}1/8$ in Fig. \ref{fig:zoom-out-8687} show the zoom-out comparison of 3DGS, Mip-Splatting and ours.
The cropped wheel shown in Fig. \ref{fig:zoom-out-8687} rendered by 3DGS \cite{kerbl20233d} is dilated with thick spokes, most severe at ${\times}1/8$, and filters in Mip-Splatting only mitigate the dilation to some extent. The result after our adaptation achieves the best visual effects and improves PSNR by 5dB compared with 3DGS \cite{kerbl20233d}. Although Scaffold-GS \cite{lu2023scaffold} and Pixel-GS \cite{zhang2024pixel} improve the full resolution rendering quality with a revised primitive growth strategy, the lack of multi-scale deformation compromises their robustness for unseen settings.
Besides, our optimization for zoom-out converges faster than for zoom-in from Fig. \ref{fig:iter}, and shows more remarkable improvement on Blender dataset. We ascribe the discrepancies in lifting effects to the quality of pseudo-GT. As Blender dataset consists of isolated objects without background, the supervision from pseudo-GT is intensive and of high quality during the adaptation stage. In comparison, the pseudo-GT provided for Mip-NeRF 360 has limited quality to cover the correct scene information and provides weaker guidance for adaptive optimization.

\vspace{1mm}
\noindent\textbf{Zoom-in.}
The zoom-in baseline models are trained on ${\times} 1/8$ downsampled images and rendered at ${\times}2, {\times}4, {\times}8$. 
As shown in Tab.\ref{tab:zoom-in-colmap} and \ref{tab:zoom-in-blender}, our method improves the zoom-in rendering quality of both 3DGS \cite{kerbl20233d} and Schaffold-GS \cite{lu2023scaffold}. As one of the SOTA methods on super-resolution, SwinIR \cite{liang2021swinir} provides HR pseudo-GT with abundant details for zoom-in deformation. In Fig. \ref{fig:zoom-in-mic}, zoom-in rendering of 3DGS exhibits needle-like aliasing due to intrinsic shrinkage bias, and Mip-Splatting \cite{yu2023mip} faces a high-frequency trade-off. Our method adjusts the Gaussians into suitable pixel coverage and generates novel views with high fidelity. 

\subsection{Ablation Study}
\label{subsec: ablation_study}
\begin{figure}[h!]
\vspace{-0.4cm}
    \begin{subfigure}[b]{0.49\textwidth}
    \centering
    \includegraphics[trim=6.5cm 7cm 5.5cm 7cm,clip=true,width=\linewidth]{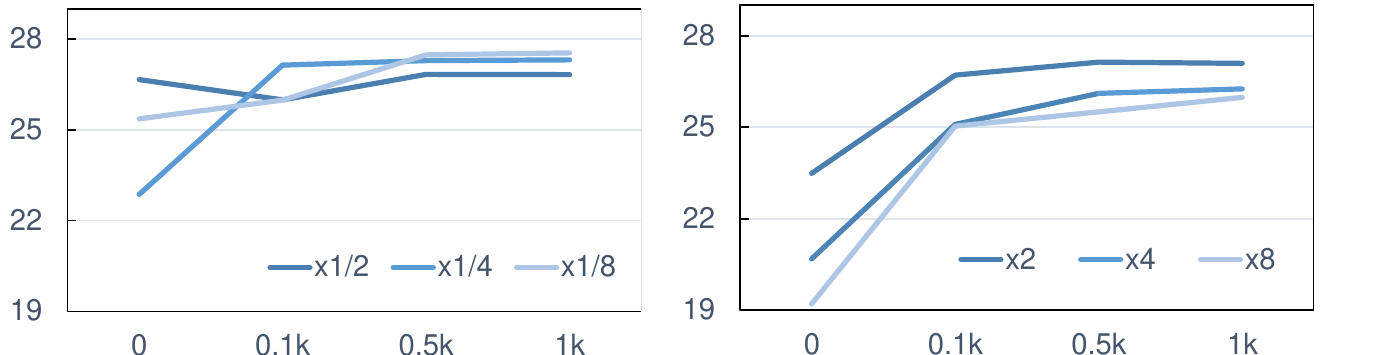}
    \vspace{-0.4cm}
    \caption{Training Iterations (PSNR in dB).}
    \label{fig:iter}
    \end{subfigure}
    \begin{subfigure}{0.47\textwidth}
    \centering
    \includegraphics[trim=5cm 7.4cm 6cm 6cm,clip=true,width=\linewidth]{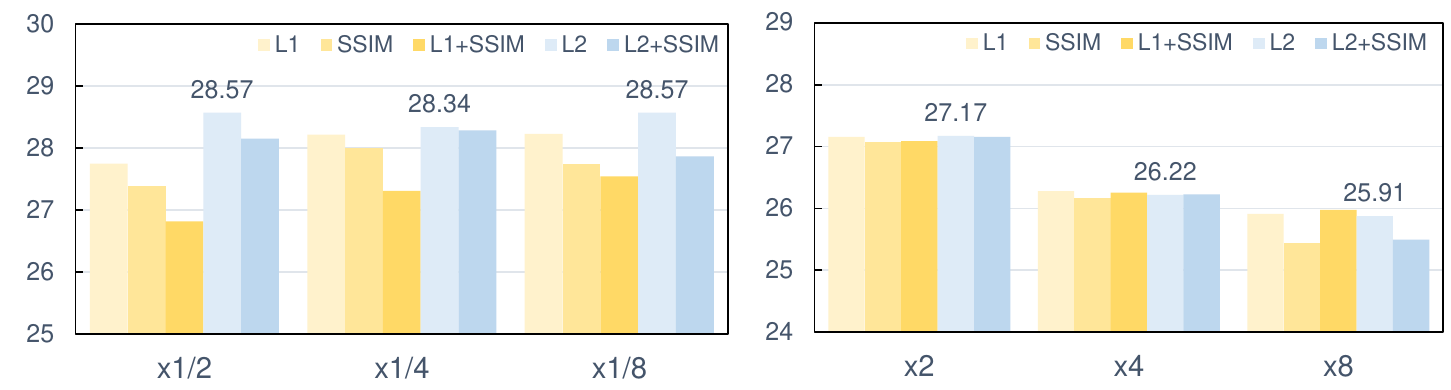}
    \vspace{-0.5cm}
    \caption{Loss Functions (PSNR in dB).}
    \label{fig:loss-func}
    \end{subfigure}
    \vspace{-0.4cm}
    \caption{Ablation on training iterations and loss functions (Mip-NeRF 360 dataset \cite{barron2022mip}).}
    \vspace{-0.4cm}
\end{figure}

\vspace{1mm}
\noindent\textbf{Training Iterations.}
Starting from trained Gaussians, our optimization needs much less training time to converge. 
As shown in Fig. \ref{fig:iter}, 
our method converges within 0.5K iterations for zoom-out and 1K iterations for zoom-in, both on the order of seconds, see Tab. \ref{tab:time}. 
In comparison, the filter update in Mip-Splatting \cite{yu2023mip} needs more time and is not compatible with 3DGS variations, e.g. \cite{lu2023scaffold}.

\vspace{1mm}
\noindent\textbf{Loss Functions.}
Originally, 3DGS takes SSIM and L1-loss weighted at 0.8 and 0.2 for training \cite{kerbl20233d}, while it is simplified to L2 loss in our optimization. We illustrate the metrics using different loss functions in Fig. \ref{fig:loss-func}. 
\begin{table}[h!]
\vspace{-0.4cm}
    \centering
    \begin{tabular}{lcc}
        \hline
        Method&\textbf{Zoom-out} &\textbf{Zoom-in} \\ 
        \hline
        3DGS \cite{kerbl20233d} & 2h30m &	8m \\
        3DGS-Ours & +40s & +75s  \\
        Scaffold-GS \cite{lu2023scaffold} & 3h7m & 12m \\
        Mip-Splatting \cite{yu2023mip}& 2h40m & 8m \\
        \hline
    \end{tabular}
    \vspace{-0.2cm}
    \caption{Comparison of training time on bicycle scene. (Ours: average time over three scales.)}
    \label{tab:time}
    \vspace{-0.5cm}
\end{table}
\begin{table}[!ht]
    \centering
    \vspace{-0.2cm}
    \begin{tabular}{lccc||c}
        \hline
        PSNR (dB) & Ours+test & +train & +syn.& 3DGS \cite{kerbl20233d}\\
        \hline
         Zoom-out& 30.84 & 29.80 &30.24& 22.60  \\
        Zoom-in & 24.23 & 22.01&22.67 & 19.36  \\
        \hline
    \end{tabular}
    \vspace{-0.2cm}
    \caption{Comparison of optimization views on bicycle scene.}
    \label{tab:views}
    \vspace{-0.5cm}
\end{table}

\vspace{1mm}
\noindent\textbf{Optimization Views.}
As shown in Tab. \ref{tab:views}, the performance drops when using training views (+train) because it focuses on known-view fitting instead of novel views.
We choose test views to retain the original novel-view performance and further improve the zooming quality. 
As a test-time learning strategy, our method also works with synthetic views sampled from the estimated camera trajectory. Here we report optimization results with 50 synthetic views (+syn.)


\subsection{Further Study}
\begin{figure}[h!]
    \centering
    \vspace{-0.3cm}
        \includegraphics[trim=4.3cm 6.7cm 3cm 5cm,clip=true,width=1.05\linewidth]{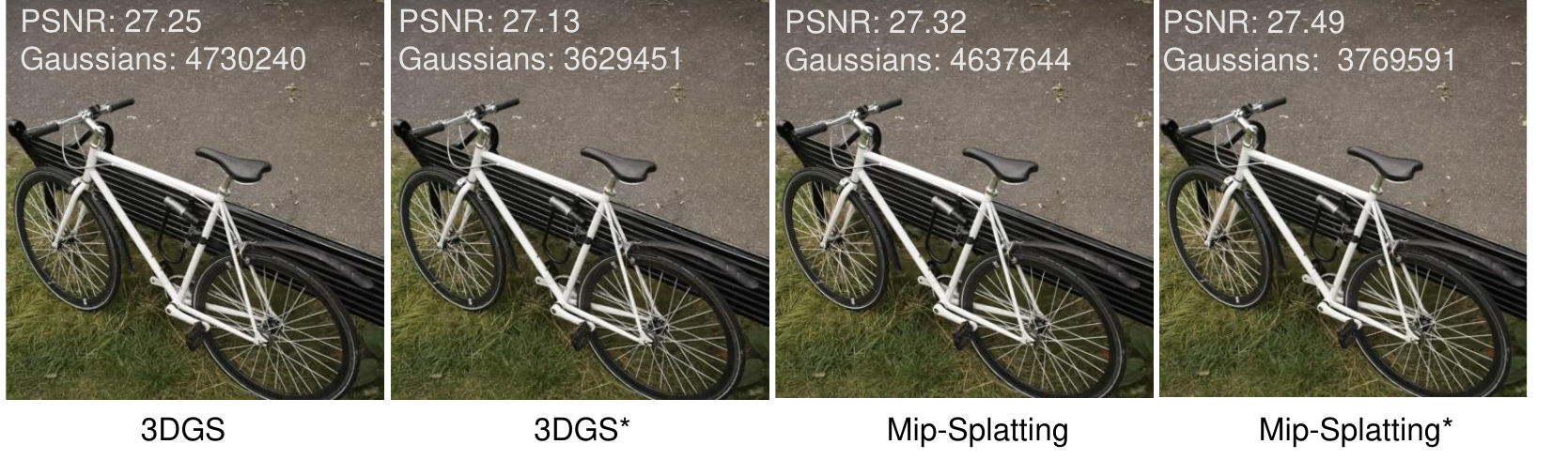}
        \vspace{-0.8cm}
        \caption{Results of selective rendering on bicycle scene trained at $\times 1/8$. (*: selective rendering with opacity$>0.01$)}
        \label{fig:redundant}
        \vspace{-0.3cm}
\end{figure}

\vspace{1mm}
\noindent\textbf{Number of Primitives.}
We find that decreasing the number of splatted Gaussians in a certain range doesn't influence the rendering performance.
Given a trained 3DGS model, rendering merely high-opacity primitives leads to comparable performance while up to $25\%$ reduction on the number of splatted primitives in Fig. \ref{fig:redundant}. In Mip-Splatting \cite{yu2023mip}, fewer primitives yield even slightly better results, which further proves the redundancy. Based on these analyses, we keep active pruning throughout optimization.

\begin{figure}[h!]
\vspace{-0.3cm}
    \centering
    \includegraphics[trim=8.4cm 9cm 8.5cm 7.5cm,clip=true,width=1.1\linewidth]{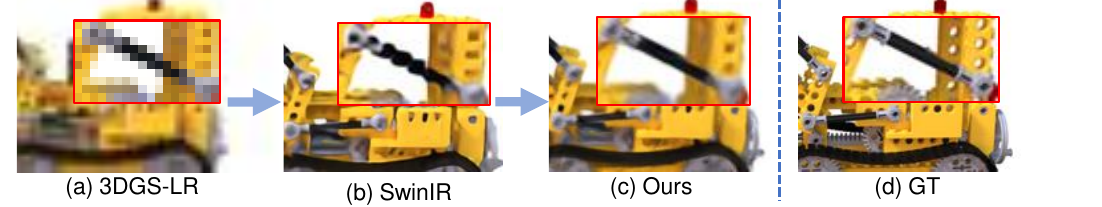}
    \vspace{-0.8cm}
    \caption{Our method refines view-consistency in 3D model.}
    \label{fig:false_label}
    \vspace{-0.4cm}
\end{figure}

\vspace{1mm}
\noindent\textbf{View Consistency.}
Although we adopt SwinIR \cite{liang2021swinir} to construct mipmap independently for each image without scene-level fine-tuning, the injected appearance is further refined within the 3D model for view-consistency, see Fig. \ref{fig:false_label}.

\begin{figure}[h!]
\vspace{-0.2cm}
    \centering
    \includegraphics[trim=8.5cm 9cm 8.5cm 7.5cm,clip=true,width=\linewidth]{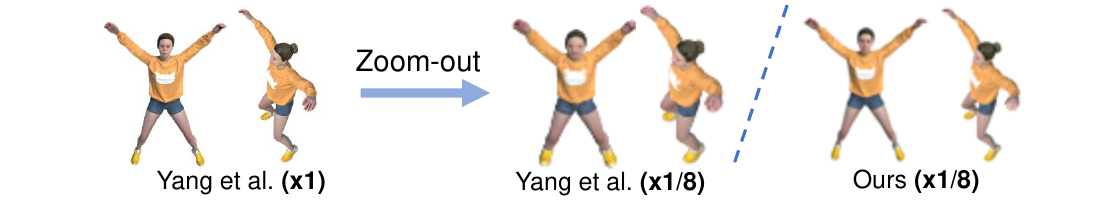}
    \vspace{-0.8cm}
    \caption{Our method is applicable in dynamic scenes, e.g. to solve zoom-out dilation in \cite{yang2023deformable}.}
    \label{fig:deform}
    \vspace{-0.55cm}
\end{figure}

\vspace{1mm}
\noindent\textbf{Application in Dynamic Scenes.}
Our method is applicable in dynamic 3DGS models for anti-aliasing rendering. Fig. \ref{fig:deform} shows our plug-in effect in \cite{yang2023deformable} with no disturbance on the deformation network. 


\section{Conclusion}

We present a unified method to fit scale-adaptive Gaussians for alias-free novel view synthesis. 
Given an unseen scale at test-time,
the designed mipmap-like pseudo-GT allows adaptive adjustment of 3D Gaussians consistent with the zoom factor.
Our method involves minimal modification towards the 3DGS pipeline and is applicable for any pre-trained Gaussian Splatting model as a plug-in module to mitigate aliasing for out-of-distribution generalization. Finally, our method converges on the order of seconds and helps to remove primitive redundancy, maintaining the fine scene representation without sacrificing real-time efficiency. 
{
    \small
    \bibliographystyle{ieeenat_fullname}
    \bibliography{main}
}


\end{document}